\documentclass[preprint,12pt]{elsarticle}

\usepackage{amssymb}

\usepackage{amsmath}
\usepackage{booktabs}
\usepackage{graphicx}
\usepackage{subcaption}
\usepackage{bm}

\newcommand{\R}{{\mathbb{R}}}

\newcommand{\Rn}{{\mathbb{R}}^{n}}

\newcommand{\RRm}{{\mathbb{R}^{2m}}}

\newcommand{\Rnn}{{\mathbb{R}^{n \times n}}}

\newcommand{\EE}{\mathbb{E}}

\newcommand{\T}{{\text{T}}}     

\newcommand{\pdot}{{\Dot{p}}}
\newcommand{\qdot}{{\Dot{q}}}

\newcommand{\xdot}{{\Dot{x}}}

\newcommand{\eye}{{\boldsymbol{I}}}
\newcommand{\zero}{{\boldsymbol{0}}}

\newcommand{\RKHS}{\mathcal{H}_{K}}

\newcommand{\odd}{\text{odd}}

\newcommand{\beq}{\begin{equation}}
\newcommand{\eeq}{\end{equation}}
\newcommand{\ba}{\left[\begin{array}}
\newcommand{\ea}{\end{array}\right]}
\newcommand{\bb}{\begin{bmatrix}}
\newcommand{\eb}{\end{bmatrix}}

\newcommand{\la}{\left\langle}
\newcommand{\ra}{\right\rangle}

\DeclareMathOperator*{\argmin}{\arg\min}
\DeclareMathOperator{\spn}{span}

\newcommand{\hboks}{\hfill$\square$\break}

\newcommand{\inftyint}{\int_{-\infty}^{\infty}}

\newcommand{\boldfhat}{\hat{\bm{f}}}

\newcommand{\boldpdot}{\Dot{\bm{p}}}
\newcommand{\boldqdot}{\Dot{\bm{q}}}

\newcommand{\boldxhat}{\hat{\bm{x}}}

\newcommand{\boldxdot}{\Dot{\bm{x}}}

\newcommand{\boldalpha}{\bm{\alpha}}

\newcommand{\boldnabla}{\bm{\nabla}}

\newcommand{\boldphi}{\bm{\phi}}

\newcommand{\thetadot}{\Dot{\theta}}

\newcommand{\thetaddot}{\Ddot{\theta}}

\newcommand{\boldGamma}{\bm{\Gamma}}

\newcommand{\boldPhi}{\bm{\Phi}}
\newcommand{\boldPsi}{\bm{\Psi}}

\newcommand{\bolda}{{\bm{a}}}

\newcommand{\boldc}{{\bm{c}}}

\newcommand{\boldf}{{\bm{f}}}
\newcommand{\boldg}{{\bm{g}}}

\newcommand{\boldp}{{\bm{p}}}
\newcommand{\boldq}{{\bm{q}}}
\newcommand{\boldr}{{\bm{r}}}

\newcommand{\boldv}{{\bm{v}}}
\newcommand{\boldw}{{\bm{w}}}
\newcommand{\boldx}{{\bm{x}}}
\newcommand{\boldy}{{\bm{y}}}
\newcommand{\boldz}{{\bm{z}}}

\newcommand{\boldB}{{\bm{B}}}

\newcommand{\boldG}{{\bm{G}}}

\newcommand{\boldJ}{{\bm{J}}}
\newcommand{\boldK}{{\bm{K}}}

\newcommand{\calF}{{\mathcal{F}}}

\newcommand{\calH}{{\mathcal{H}}}

\newcommand{\calN}{{\mathcal{N}}}

\newcommand{\calS}{{\mathcal{S}}}

\newcommand{\calZ}{{\mathcal{Z}}}

\newcommand{\mbbE}{{\mathbb{E}}}

\usepackage{xcolor}
\usepackage{soul}

\journal{European Journal of Control}

\begin{document}

\begin{frontmatter}




\title{Learning Hamiltonian Dynamics with Reproducing Kernel Hilbert Spaces and Random Features}


\author[aff1]{Torbjørn Smith\corref{cor1}}
\ead{torbjorn.smith@ntnu.no}

\author[aff1]{Olav Egeland}
\ead{olav.egeland@ntnu.no}

\affiliation[aff1]{
    organization    = {Department of Mechanical and Industrial Engineering, Norwegian University of Science and Technology (NTNU)},
    addressline     = {Richard Birkelands vei 2B}, 
    city            = {Trondheim},
    postcode        = {7491},
    country         = {Norway}
}

\cortext[cor1]{Corresponding author.}


\begin{abstract}
A method for learning Hamiltonian dynamics from a limited and noisy dataset is proposed. The method learns a Hamiltonian vector field on a reproducing kernel Hilbert space (RKHS) of inherently Hamiltonian vector fields, and in particular, odd Hamiltonian vector fields. This is done with a symplectic kernel, and it is shown how the kernel can be modified to an odd symplectic kernel to impose the odd symmetry. A random feature approximation is developed for the proposed odd kernel to reduce the problem size. The performance of the method is validated in simulations for three Hamiltonian systems. It is demonstrated that the use of an odd symplectic kernel improves prediction accuracy and data efficiency, and that the learned vector fields are Hamiltonian and exhibit the imposed odd symmetry characteristics.
\end{abstract}







\end{frontmatter}


\section{Introduction}\label{sec:1_introduction}

Learning of dynamical systems is an important area of research in robotics and control engineering, and data-driven methods have emerged as a robust approach for system identification, where classical analytical methods may be impractical. The aim is to utilize machine learning to derive a model of the underlying dynamical system from a set of measurement data \cite{Brunton2022}. The efficacy of data-driven methods depends on the quality of the training dataset. Notably, these methods may encounter challenges such as limited generalization beyond the provided dataset \cite{Sindhwani2018} and may be susceptible to overfitting in cases where the data set is limited or noisy \cite{Ahmadi2020}. Assembling a viable dataset may also be labor intensive or even impractical in real-world or online applications. Furthermore, as the data set grows, the computational cost of learning the model increases, as does the inference time of the final learned model for some learning methods \cite{Sindhwani2018}\cite{Singh2021}. It is important that the learned models are stable and robust, particularly for safety-critical control applications \cite{Revay2020}. In response to these challenges, researchers have developed different strategies to guide or restrict the learning of dynamical systems using prior information, which can lead to satisfactory results even with limited datasets.

There may be physical laws or contextual knowledge about the system that is insufficient to derive analytical models, but that may be used to improve the learning of dynamical systems. The mathematical formalism for learning dynamical systems with side information was presented in \cite{Ahmadi2020}, where a range of side constraints were outlined and demonstrated. An important physical law is energy conservation, which can be enforced through the use of the Hamiltonian formalism \cite{Greydanus2019}. By learning dynamical systems with the Hamiltonian formalism, the learned system is constrained to conserve the total energy of the system in the phase space. This has proved successful in improving accuracy and generalizability in several publications \cite{Greydanus2019}\cite{Ahmadi2018}\cite{Zhong2020}\cite{Chen2020}, and is of high relevancy from a control perspective \citep{Ahmadi2018}. Contextual knowledge, such as symmetry, can also be enforced through side constraints \cite{Ahmadi2020}, and a wide range of physical systems commonly seen in the data-driven modeling literature, such as the harmonic oscillator, pendulum, cart-pole, and acrobot are odd symmetric. Enforcing symmetry in the learning of dynamical systems improves the generalizability of the model and is particularly useful for instances where the data set is limited to a subset of the domain for which the learned model is to be applied \cite{Espinoza2005}. Enforcing odd and even symmetry proved useful for learning more accurate and generalizable models for price prediction \cite{Krejnik2012}, mechanical systems \cite{Ahmadi2020}, and chaotic systems such as the Lorenz attractor \cite{Espinoza2005}\cite{Aguirre2004}.

\subsection{Contribution}

In this paper, we show how Hamiltonian dynamical systems with odd vector fields can be learned in a reproducing kernel Hilbert space (RKHS) by developing a kernel that ensures that the learned vector fields are odd Hamiltonian vector fields. The proposed kernel is approximated using random Fourier features (RFF) for dimensionality reduction. We also include a novel approximation of the odd and even kernels using RFF. Encoding the constraints in the kernel reduces the learning time as the straightforward closed-form solution of the learning problem is retained. Three illustrative simulation examples demonstrate that the generalization properties of the learned model for out-of-sample data points, which are points that are outside of the region of the training points, are improved through the additional constraints and that energy preservation and odd symmetry are encoded in the final model.

A preliminary version of the proposed learning algorithm was presented in \cite{Smith2024}. In the present paper, we extend the method in \cite{Smith2024} by incorporating RFF to approximate the proposed kernel. Furthermore, the simulation experiments are expanded to encompass more sophisticated Hamiltonian systems that are common in the system identification literature.

\subsection{Related work}

In the following related work on data-driven modeling and the learning of dynamical systems with kernels is presented with an emphasis on work that includes constrained learning. The relevant work related to learning Hamiltonian dynamical systems is also presented.

\subsubsection{Data-driven modeling with kernels}

In \cite{Krejnik2012}, financial price prediction was explored using a data-driven approach with functions in an RKHS. By designing an odd reproducing kernel that imposed an odd symmetry constraint on the price action, the learned model demonstrated improved prediction accuracy and reduced overfitting compared to the unconstrained method.

The \textit{learning-from-demonstrations} problem was addressed in  \cite{Sindhwani2018}, where the focus was on copying human demonstrations using a data-driven approach. A dynamical system was learned in an RKHS with RFF for dimensionality reduction, allowing the imitation of human-drawn shapes. The learned dynamical system included desired equilibrium points, and point-wise contraction constraints were enforced along the trajectory to create a contraction region around the desired path, conditioning the learned vector field. Learning nonlinear dynamics with a stabilizability constraint was investigated in \cite{Singh2021}. The dynamics were learned using a contraction constraint, and the model was evaluated using a planar drone. The method enhanced trajectory generation, tracking, and data efficiency. The model was learned in an RKHS, and utilized RFF for dimensionality reduction. In \cite{Khosravi2021}, they performed nonlinear system identification by incorporating constraints enforcing prior knowledge of the region of attraction. The stability region was enforced using a Lyapunov function, and the hypothesis space for the learned model was an RKHS. \cite{Thorpe2023} explores learning dynamical systems in an RKHS, incorporating a bias term in the regularized least squares cost to embed prior knowledge, improving data efficiency and out-of-sample generalization. In \cite{VanWaarde2023} the identification of nonlinear input-output operators in an RKHS is studied. Nonexpansive operators are introduced to identify operators that satisfy a wide range of dissipativity and integral quadratic constraints.

\subsubsection{Learning Hamiltonian dynamics}

Polynomial basis functions were used in \cite{Ahmadi2018} to investigate control-oriented learning of Lagrangian and Hamiltonian systems. It was demonstrated that accurate and generalized learning from a limited number of trajectories could be achieved by learning these functions.

The work in \cite{Greydanus2019} focused on learning the Hamiltonian dynamics of energy-conserving systems using neural networks, where the Hamiltonian was learned as a parametric function. This approach significantly enhanced the predictive accuracy of the learned system. Building upon this, \cite{Zhong2020} further refined the method by eliminating the need for higher-order derivatives of the generalized coordinate and incorporating the option for energy-based control. In \cite{Chen2020}, the work in \cite{Greydanus2019} was extended by using the symplectic Leapfrog integrator to integrate the partial derivatives of the learned Hamiltonian. The loss was then back-propagated through the integrator over multiple time steps, resulting in improved learning performance for more complex and noisy Hamiltonian systems.

System identification of Hamiltonian vector fields has also been conducted using Gaussian process (GP) models. The advantage of GP modeling is that uncertainty in the dataset is considered at the cost of computational complexity {\cite{Rasmussen2006}}. In \cite{Rath2021}, the symplectic Gaussian process regression (SympGPR) method was presented. The method utilized Hamiltonian mechanics to derive the covariance function in the GP model for learning energy-conserving or Hamiltonian vector fields from trajectory and derivative data. The Hamiltonian function was modeled using a single output GP, and the covariance function was derived by taking the symplectic gradient of the Hamiltonian function. SympGPR was further developed in \cite{Tanaka2022} with the introduction of Symplectic Spectrum Gaussian Processes (SSGPs), which allowed for learning both energy-preserving and dissipative Hamiltonian vector fields. The need for derivative data was eliminated by approximating the GP prior with symplectic structure preserving random Fourier features. This also allowed for more efficient sampling of the learned vector field. SSGP was compared to several existing methods, and it was shown in numerical experiments that SSGP was among the most accurate and data-efficient, especially for longer prediction horizons. Both methods utilize the symplectic structure to enforce energy conservation, but neither method includes symmetry as a side constraint.

\subsection{Paper structure}

The paper is organized as follows: Section \ref{sec:2_problem_formulation} outlines the problem addressed in this work. Section \ref{sec:3_preliminaries} provides a review of the relevant theory related to reproducing kernel Hilbert spaces, regularized least-squares, random Fourier features, and Hamiltonian mechanics. The main contribution is detailed in Section \ref{sec:4_method}, where the random feature approximation for odd reproducing kernels is presented. This includes the odd symplectic kernel, which is approximated using random features. Section \ref{sec:5_experiments} presents the numerical simulation experiments used to validate the proposed method. Finally, Section \ref{sec:6_conclusion} presents the conclusion and future work.
\section{Problem formulation}\label{sec:2_problem_formulation}

This paper explores learning the dynamics of an unknown system from limited data. The system dynamics are given by the vector field
\beq\label{vector_field_dynamic_model}
    \boldxdot = \boldf (\boldx)
\eeq
where ${\boldx \in \Rn}$ is the state vector, ${\boldxdot \in \Rn}$ is the time derivative of the state vector, and ${\boldf : \Rn \rightarrow \Rn}$ are the system dynamics. It is assumed that $\boldy=\boldxdot$ is available as a measurement or from numerical differentiation. Given a set of ${N}$ data points ${ \{ (\boldx_{i}, \boldy_{i}) \in \Rn \times \Rn \}_{i=1}^{N}}$ from simulations or measurements, the aim is to learn a function ${\boldfhat \in \calF}$, where $\calF$ is a class of functions. The class of functions $\calF$ will be the functions of a reproducing kernel Hilbert space (RKHS) determined by a reproducing kernel \cite{Aronszajn1950}. The function $\boldfhat$ is found by the regularized minimization problem \cite{Micchelli2005}
\beq\label{eq:basic_learning_problem}
    \boldfhat = \argmin_{\boldf \in \calF} \frac{1}{N} \sum_{i=1}^{N} \| \boldf(\boldx_i) - \boldy_i \|^{2}
    + \lambda\| \boldf \|^2_{\calF}
\eeq
where ${\lambda > 0}$ is the regularization parameter. The least-squares optimization in combination with the regularization ensures that noisy data, like uncertainty in $\dot\boldx$, will be handled well, and that a tendency in overfitting is limited by the regularization term. A special property of the solution of \eqref{eq:basic_learning_problem} with RKHS techniques is that if the function $\hat \boldf$ converges to $\boldf$ in the norm of the RKHS $\calF$, then the function value $\hat\boldf(\boldx)$ converges to $\boldf(\boldx)$ in the norm of $\Rn$ for every $\boldx$ (see Section~{\ref{seq:RKHS}}).

It is well known that this approach may lead to inaccurate generalization beyond the data set used to learn the dynamical model. Furthermore, if the trajectories in the data set are limited and noisy, the learned dynamical model may fail to capture the dynamics of the underlying system due to overfitting.

It is assumed that there is some information about the physical properties of the dynamical system. This type of side information about the system was treated in \cite{Ahmadi2020} where the function class $\calF$ was polynomial functions. The additional information about the dynamics was included as side constraints in \cite{Ahmadi2020} by defining a subset $\calS_i \subset \calF$ for each side constraint $i$, so that the function $\boldfhat$ satisfies the side constraint whenever $\boldfhat\in\calS_i$. The learning problem including the side constraints can be formulated as 
\beq\label{eq:learning_problem_with_side_info}
    \boldfhat = \argmin_{\boldf \in \calF \cap \calS_{1} \cap \dots \cap \calS_{k}} \frac{1}{N} \sum_{i=1}^{N} \| \boldf(\boldx_i) - \boldy_i \|^{2}
\eeq
In this paper the side constraints are instead handled by defining a reproducing kernel which ensures that the RKHS function class $\calF$ inherently satisfies the relevant side constraints. It is well-known that this can be done to have a RKHS where the vector field $\boldfhat$ is curl-free or divergence-free \cite{Brault2016}, symplectic \cite{Boffi2022}, odd or even \cite{Krejnik2012}. It is also possible to impose additional side constraints like contraction \cite{Sindhwani2018} or stabilizability \cite{Singh2021} along the trajectories of the dataset, but this will not be addressed in this paper. 

In this paper the function class $\calF$ will be a reproducing kernel Hilbert space (RKHS). The side constraints are that the state dynamics are symplectic, and, in addition, odd in the sense that $\boldf(-\boldx) = -\boldf(\boldx)$, and this is ensured by selecting an appropriate reproducing kernel. 
\section{Preliminaries}\label{sec:3_preliminaries}

\subsection{Reproducing kernel Hilbert space}
\label{seq:RKHS}
The theory for reproducing kernel Hilbert spaces (RKHS) was formulated by Aronszajn in \cite{Aronszajn1950}. This was extended to vector-valued functions in \cite{Micchelli2005} and \cite{Carmeli2010}. This theory will be used in this paper for vector fields $\boldf:\R^n\rightarrow\R^n$. 

A map $\boldK:\R^n\times\R^n \rightarrow \R^{n\times n}$ is called a vector-valued reproducing kernel if for any $N>0$ and any sets $\boldx_1,\ldots,\boldx_N \in\R^n$ and $\boldy_1,\ldots,\boldy_N \in\R^n$, the kernel is positive definite in the sense that 
\beq\label{RKHS_Vector_Micchelli_Positive_definite_kernel}
\sum_{i=1}^N\sum_{j=1}^N \left\langle \boldy_j,\boldK(\boldx_j,\boldx_i)\boldy_i \right\rangle_{\R^n} \geq 0
\eeq
Let the map $\boldK_x\boldy:\R^n \rightarrow\R^n$ be defined for every $\boldx,\boldz,\boldy\in\R^n$ by
\beq\label{def-kx}
(\boldK_x\boldy)(\boldz) = \boldK(\boldz,\boldx)\boldy
\eeq
The notation $\boldK(\cdot,\boldx) = \boldK_x$ is also used. Let $\calH_K$ be a Hilbert space of functions $\boldf: \R^n \rightarrow \R^n$ with inner product $\la \cdot, \cdot\ra_{\calH_K}$. Then $\calH_K$ is the reproducing kernel Hilbert space (RKHS) corresponding to the reproducing kernel $\boldK$ if for all $\boldx,\boldy\in\R^n$
\beq
\boldK_x\boldy \in \calH_K 
\eeq
and
\beq\label{Reproducing-property}
\la \boldy,\boldf(\boldx)\ra_{\R^n}
    = \la \boldK_x\boldy,\boldf\ra_{\calH_K}
\eeq
where \eqref{Reproducing-property} is referred to as the reproducing property. Moreover, 
\begin{align}
    &\calH_K = \overline{\spn}\{\boldK_x\boldy \ |\ \forall\boldx\in\R^n, \forall\boldy\in\R^n\}
\end{align}

An important property is that \cite{Micchelli2005}
\begin{equation}\label{eq:norm-f(x)-bounded-by-norm_of-f}
    \| \boldf(\boldx) \|_{\R^n} \leq \sqrt{\| \boldK(\boldx,\boldx) \|} \| \boldf \|_{\RKHS}
\end{equation}
which implies 
\begin{equation}
    \| \boldf(\boldx) - \boldg(\boldx) \|_{\R^n} \leq \sqrt{\| \boldK(\boldx,\boldx) \|} \| \boldf - \boldg \|_{\RKHS}
\end{equation}
This means that if ${\| \boldf -\boldg\|_{\RKHS}}$ converges to zero, then ${\|\boldf(\boldx) - \boldg(\boldx)}\|_{\R^n}$ converges to zero for each ${\boldx}$.

A feature map $\boldPhi_K$ is a mapping which satisfies
\beq
\boldK(\boldx,\boldz) = \boldPhi_K(\boldx)^*\boldPhi_K(\boldz)
\eeq
The reproducing property \eqref{Reproducing-property} with $\boldf = \boldK_z\boldw\in \calH_K$ gives
\beq
\langle \boldy,(\boldK_z\boldw)(\boldx)\rangle_{\R^n} 
= \langle \boldK_x\boldy,\boldK_z\boldw\rangle_{\calH_K} 
= \langle \boldy,\boldK_x^*\boldK_z\boldw\rangle_{\R^n}
\eeq
where $\boldK_x^*$ is the adjoint of $\boldK_x$. It follows from \eqref{def-kx} that the kernel can be written
\beq
\boldK(\boldx,\boldz) = \boldK^{*}_{x}\boldK_{z}
\eeq
This means that a possible feature map is $\boldPhi_K(\boldx)=\boldK_x$, which is referred to as the canonical feature map in \cite{Minh2016}. 

A kernel is called shift-invariant if $\boldK(\boldx,\boldz) = \boldG(\boldx-\boldz)$, where $\boldG$ is called the signature of the kernel. 

\subsection{Regularized least-squares}

Consider the regularized least-squares solution
\begin{equation}\label{eq:vector_valued_regular_least_squares}
    \hat\boldf = 
    \argmin_{\boldsymbol{f} \in \mathcal{H}_{K}} \frac{1}{N} \sum_{i=1}^{N} \| \boldsymbol{f}(\boldsymbol{x}_{i}) - \boldsymbol{y}_{i} \|^{2} + \lambda \| \boldsymbol{f} \|_{\mathcal{H}_{K}}^{2}
\end{equation}
where a data set ${ \{ (\boldx_{i}, \boldy_{i}) \in \Rn \times \Rn \}_{i=1}^{N}}$ is given, $\lambda > 0$ is the regularization parameter and $\|\boldf\|_{\mathcal{H}_{K}}^{2} = \la \boldf,\boldf\ra_{\RKHS}$. The optimal solution is then
 \cite{Micchelli2005}
\begin{equation}\label{RLS-optimal-solution}
    \hat\boldf(\boldx) = \sum_{i=1}^{N} \boldK(\boldx,\boldx_i) \boldsymbol{a}_i \in \R^n
\end{equation}
where the coefficients vectors ${\boldsymbol{a}_i \in \mathbb{R}^n}$ are the unique solutions of 

\beq\label{eq:vector_valued_regular_least_squares_a_equation}
(\tilde\boldK + N\lambda \eye_{Nn})\tilde\bolda = \tilde\boldy
\eeq
where 
\beq
\tilde\boldK = \ba{ccc} 
 \boldK(\boldx_1,\boldx_1) & \ldots & \boldK(\boldx_1,\boldx_N) \\
\vdots & \ddots & \vdots \\
\boldK(\boldx_N,\boldx_1) & \ldots & \boldK(\boldx_N,\boldx_N)
\ea \in \R^{Nn\times Nn} 
\eeq
is the Gram matrix, $\tilde\bolda = [\bolda_1^\T,\ldots,\bolda_N^\T]^\T$ and $\tilde \boldy = [\boldy_1^\T,\ldots,\boldy_N^\T]^\T$.

\subsection{Random Fourier features}

Random Fourier features (RFF) were introduced for scalar-valued functions in \cite{Rahimi2007} where Bochner's theorem and the inverse Fourier transform were used to generate random features that could be used to approximate a shift-invariant scalar kernel. This was extended to vector-valued shift-invariant kernels in \cite{Brault2016} and \cite{Minh2016}. The motivation for using RFF is a significant reduction of the computational complexity in the solution of \eqref{eq:vector_valued_regular_least_squares}.

\textbf{Assumption 1:}  {\em Given the shift-invariant reproducing kernel $\boldK(\boldx,\boldz) = \boldG(\boldx-\boldz) \in \R^{n\times n}$ with signature $\boldG$, there is a  probability density function $p(\boldw)$ and a matrix  $\boldB (\boldw) \in\R^{n\times n_1}$ where $n_1\leq n$ so that
\beq
\boldG(\boldx) 
= \int_{\R^n} \cos(\boldx^\T\boldw) \boldB(\boldw)\boldB(\boldw)^\T p(\boldw)d\boldw
\eeq
where $\boldw\in\R^n$. 
}

It is noted that under Assumption 1 the signature is the expected value
\beq
\boldG(\boldx) = \EE_w\left[  \cos(\boldx^\T\boldw) \boldB(\boldw)\boldB(\boldw)^\T  \right]
\eeq
where $\boldw \sim p(\boldw)$. This leads to
\begin{align}
\boldG(\boldx-\boldz) 
&= \EE_w\left[ \tilde\boldPhi(\boldx,\boldw)^\T\tilde\boldPhi(\boldz,\boldw) \right]
\end{align}
where \cite{Brault2016} 
\beq
\tilde\boldPhi(\boldx,\boldw) 
= \left[\begin{array}{c}
\cos(\boldx^\T\boldw)\boldB(\boldw)^\T\\
\sin(\boldx^\T\boldw)\boldB(\boldw)^\T 
\end{array}
\right] \in \R^{2n_1\times n}
\eeq
An approximation of the kernel in terms of the random Fourier features  $\boldPsi(\boldx)$ is then given by 
\beq\label{RFF-K-general-case}
\boldK(\boldx,\boldz) \approx \boldPsi(\boldx)^{\T} \boldPsi(\boldz)
\eeq
where 
\begin{equation}\label{RFF-general-case}
    \boldPsi(\boldx) = \frac{1}{\sqrt{d}} 
    \begin{bmatrix}
        \cos(\boldw_{1}^{\T}\boldx)\boldB(\boldw_{1})^{\T}\\
        \vdots\\
        \cos(\boldw_{d}^{\T}\boldx)\boldB(\boldw_{d})^{\T}\\
        \sin(\boldw_{1}^{\T}\boldx)\boldB(\boldw_{1})^{\T}\\
        \vdots\\
        \sin(\boldw_{d}^{\T}\boldx)\boldB(\boldw_{d})^{\T}
    \end{bmatrix}
    \in \R^{2dn_1\times n}
\end{equation}
and $\boldw_1,\ldots,\boldw_d$ are drawn with distribution $p(\boldw)$.

If an RFF approximation \eqref{RFF-K-general-case} of the kernel is used to solve the regularized least-squares problem \eqref{eq:vector_valued_regular_least_squares}, then insertion of \eqref{RFF-K-general-case} in \eqref{RLS-optimal-solution} gives
\begin{equation}\label{function-value-RRF-approximation}
    \hat\boldf(\boldx) = \boldPsi(\boldx)^{\T} \boldsymbol{\alpha}
\end{equation}
where the coefficient vector is $\boldsymbol{\alpha} = \sum_{i=1}^{N} \boldPsi(\boldx_i) \bolda_i \in \R^{2dn_1}$. The vector $\boldsymbol{\alpha}$ which optimizes \eqref{eq:vector_valued_regular_least_squares} is computed by solving the linear equation  \begin{equation}\label{eq:rff_vector_valued_regular_least_squares_minimum}
 \left( \sum_{i=1}^N \left(\boldPsi(\boldx_i)\boldPsi(\boldx_i)^{\T} + \lambda \eye_{2dn_1}\right) \right) \boldalpha = \sum_{i=1}^N \boldPsi(\boldx_i) \boldy_i
\end{equation}
This requires the solution of a linear system of dimension $2dn_1\times 2dn_1$. Since $d$ is typically selected so that $2dn_1 \ll Nn$, this solution requires significantly less computation than the original solution of \eqref{eq:vector_valued_regular_least_squares_a_equation} with dimension $Nn\times Nn$.

\subsection{RFF for Gaussian and curl-free kernels}

The kernels presented in this section satisfy Assumption 1, and the RFF approximations differ only in the definition of ${\boldB(\boldw)}$. 

The scalar shift-invariant Gaussian kernel \cite{Scholkopf2001} is a reproducing kernel given by
\beq\label{def-gaussian-kernel}
k_\sigma(\boldx,\boldz) = g_\sigma(\boldx-\boldz) 
= e^{-\frac{\|\boldx-\boldz\|^2}{2\sigma^2}}
\eeq
The RFF approximation is given by \eqref{RFF-general-case} with ${\boldB(\boldw) = 1}$ and ${\boldw \sim p_{\sigma}(\boldw) =}$ ${ \calN\left(\zero,\sigma^{-2} \eye_n\right)}$ \cite{Rahimi2007}. 

The Gaussian separable kernel \cite{Sindhwani2018} 
\beq\label{Gaussian-separable-kernel}
\boldK_\sigma(\boldx,\boldz) = k_\sigma(\boldx,\boldz) \eye_n
\eeq
where ${\eye_n}$ is the ${n\times n}$ identity matrix, is a vector-valued reproducing kernel, and the RFF approximation is given by \eqref{RFF-general-case} with ${\boldB(\boldw) = \eye_n}$, ${\boldw \sim p_{\sigma}(\boldw)}$, and $n_1 = n$. 


The curl-free kernel \cite{Fuselier2006} $\boldK_c(\boldx,\boldz) = \boldG_c(\boldx-\boldz) \in \R^{n\times n}$ is a vector valued reproducing kernel derived from the Gaussian kernel as 
\beq\label{eq:curl_free_kernel_from_gaussian}
    \boldG_{c}(\boldx) = -\boldnabla \boldnabla^{\T}g_\sigma(\boldx) = \frac{1}{\sigma^2} e^{-\frac{\boldx^{\T} \boldx}{2 \sigma^2}} \left( \eye_n - \frac{\boldx \boldx^{\T}}{\sigma^2} \right)
\eeq
where $\boldnabla = [\partial/\partial x_1,\ldots,\partial/\partial x_n]^\T$. The RFF approximation \cite{Sindhwani2018} is given by \eqref{RFF-general-case} with $\boldB(\boldw) = \boldw$, ${\boldw \sim p_{\sigma}(\boldw)}$, and $n_1 = 1$.


\subsection{Hamiltonian dynamics}

Consider a Hamiltonian system with generalized coordinates $\boldq \in \R^m$, momentum variables $\boldp \in \R^m$ and state vector ${\boldx = \bb \boldq^{\T},\boldp^{\T} \eb^{\T} \in \R^n}$ where $n=2m$. The Hamiltonian is assumed to be given as the energy $H(\boldx) = T(\boldq,\boldp) + U(\boldq)$ where $T(\boldq,\boldp)$ is the kinetic energy and $U(\boldq)$ is the potential energy. The numerical value of the Hamiltonian $H$ will depend on the definition of the zero level of the potential $U(\boldq)$. The equations of motion for the system are given by 
\begin{equation}\label{eq:Hamiltonian_dynamics_J}
    \boldxdot = \boldf(\boldx) = \boldJ \boldnabla H (\boldx)
\end{equation}
where
\beq\label{eq:symplectic_matrix}
    \boldJ = 
    \bb 
        \boldsymbol{0} & \eye_m\\
        -\eye_m  & \boldsymbol{0}
    \eb
    \in \Rnn
\eeq
is the skew-symmetric symplectic matrix. The time derivative of the Hamiltonian is 
\begin{equation}\label{eq:hamiltonian_time_derivative_zero}
    \frac{dH(\boldx)}{dt} 
    = (\boldnabla^\T H (\boldx))^\T \boldxdot 
    = (\boldnabla^\T H (\boldx))^\T \boldJ \boldnabla H (\boldx) =  0
\end{equation}
where it is used that $\boldJ$ is skew-symmetric. 

Consider the system 
\beq\label{Symplectic-system-dotx=fx}
\boldxdot = \boldf_s(\boldx)
\eeq
where ${\boldx \in \RRm}$. The flow of the system is given by $\boldphi_{t}(\boldx_0) = \boldx(t)$ where ${\boldx(t)}$ is the solution of \eqref{Symplectic-system-dotx=fx} with initial condition $\boldx(0) = \boldx_0$.

\textbf{Definition 1:}
{\em Let ${\boldPsi(t) = \partial \boldphi_{t}(\boldx_0) / \partial \boldx_{0}}$ where $\boldphi_{t}(\boldx_0)$ be the flow of \eqref{Symplectic-system-dotx=fx}. The system \eqref{Symplectic-system-dotx=fx} is said to be symplectic if 
\beq\label{eq:symplectic_condition}
    \boldPsi(t)^{\T} \boldJ \boldPsi(t) = \boldJ
\eeq
for all $t\geq 0$.} 

The system \eqref{Symplectic-system-dotx=fx} is symplectic if and only if there is a Hamiltonian $H_s(\boldx)$ so that $\boldf_s(\boldx) = \boldJ \boldnabla H_s (\boldx)$ \cite[Theorem 2.6]{Hairer2006}.

\section{Odd reproducing kernels}\label{sec:4_method}

In the following section, the main theoretical result of the paper is presented.

\subsection{Odd kernel}

The following lemma is a vector-valued version of the result in \cite{Krejnik2012}.

\textbf{Lemma 1:} {\em
Consider a shift-invariant reproducing kernel $\boldK(\boldx,\boldz) = \boldG(\boldx-\boldz)$ where $\boldG(\boldx) = \boldG(-\boldx)$. Then $\boldK(\boldx,\boldz) = \boldK(-\boldx,-\boldz)$, and
\begin{equation}\label{eq:odd_kernel}
    \boldK_{\mathrm{odd}}(\boldx,\boldz) = \frac{1}{2} (\boldK(\boldx,\boldz) - \boldK(-\boldx,\boldz)) 
\end{equation}
is a vector-valued reproducing kernel which is odd in the sense that 
\beq
\boldK_{\mathrm{odd}}(-\boldx,\boldz) = - \boldK_{\mathrm{odd}}(\boldx,\boldz)
\eeq
Any function $\boldf = \sum_{i=1}^N \boldK_{\mathrm{odd}}(\cdot,\boldx_i) \bolda_i \in \calH_{\mathrm{odd}}$ where $\calH_{\mathrm{odd}}$ is the RKHS defined by $\boldK_{\mathrm{odd}}$ will then be odd, since $\boldf(-\boldx) = -\boldf(\boldx)$. 
}

Proof:
It follows from $\boldG(\boldx) = \boldG(-\boldx)$ that
\beq
\boldK(\boldx,\boldz) = \boldG(\boldx-\boldz) = \boldG(-\boldx+\boldz) = \boldK(-\boldx,-\boldz)
\eeq
The reproducing kernel $\boldK(\boldx,\boldz)$ is positive definite, and $\boldK(-\boldx,\boldz)$ is positive definite since 
\eqref{RKHS_Vector_Micchelli_Positive_definite_kernel}
is valid for all $\boldx$. Therefore $\boldK_{\text{odd}}(\boldx,\boldz)$ is positive definite since it is the sum of two positive definite kernels. The odd property follows from
\beq
\boldK_{\mathrm{odd}}(-\boldx,\boldz) = \frac{1}{2} (\boldK(-\boldx,\boldz) - \boldK(\boldx,\boldz)) = - \boldK_{\text{odd}}(\boldx,\boldz)
\eeq
and 
\beq
\boldf(-\boldx) = \sum_{i=1}^N \boldK_{\text{odd}}(-\boldx,\boldx_i)\bolda_i
= -\sum_{i=1}^N \boldK_{\text{odd}}(\boldx,\boldx_i)\bolda_i  = -\boldf(\boldx)
\eeq
\hboks

The odd kernel will not be shift-invariant, and it will not have a signature. Instead, it is given by a difference of two signatures as
\beq\label{odd-kernel-signature-functions}
\boldK_{\text{odd}}(\boldx,\boldz) = \frac{1}{2} (\boldG(\boldx-\boldz) - \boldG(\boldx+\boldz))
\eeq
This will be used to find random Fourier features for the odd kernel.

\subsection{Random features approximation of an odd kernel}

\textbf{Proposition 1:}  {\em  Suppose that Assumption 1 holds. Then the odd kernel defined in Lemma 1 will satisfy
\begin{equation}\label{odd-kernel-expected-value}
    \boldK_{\mathrm{odd}}(\boldx,\boldz) = \mbbE_{w}\bb \sin(\boldw^{\T}\boldx) \sin(\boldw^{\T}\boldz) \boldB(\boldw)\boldB(\boldw)^\T \eb
\end{equation}
and a RFF approximation is given by 
\begin{equation}\label{odd-kernel-rff-approx}
    \boldK_{\mathrm{odd}}(\boldx,\boldz) \approx \boldPsi_o(\boldx)^{\T} \boldPsi_o(\boldz)
\end{equation}
where
\begin{equation}\label{odd-kernel-rff-approx-psi}
    \boldPsi_o(\boldx) = \frac{1}{\sqrt{d}} 
    \begin{bmatrix}
        \sin(\boldw_{1}^{\T}\boldx)\boldB(\boldw_{1})^{\T}\\
        \vdots\\
        \sin(\boldw_{d}^{\T}\boldx)\boldB(\boldw_{d})^{\T}
    \end{bmatrix}
    \in \R^{dn_1 \times n}
\end{equation}
where ${\boldw_1,\ldots, \boldw_d \in \Rn}$ are drawn with distribution ${p(\boldw)}$.
}

Proof:
Application of Bochner's theorem to the signature functions in \eqref{odd-kernel-signature-functions} gives
\begin{align}
    \boldK_{\mathrm{odd}}(\boldx,\boldz) 
    &= \frac{1}{2} \left(\boldG(\boldx - \boldz) - \boldG(\boldx + \boldz)\right) \nonumber\\
        &= \frac{1}{2} \inftyint \cos(\boldw^{\T} (\boldx - \boldz)) \boldB(\boldw)\boldB(\boldw)^\T p(\boldw) d\boldw\nonumber\\
        &\quad - \frac{1}{2} \inftyint \cos(\boldw^{\T} (\boldx + \boldz)) \boldB(\boldw)\boldB(\boldw)^\T p(\boldw) d\boldw
    \nonumber\\
    &= \inftyint \sin(\boldw^{\T}\boldx) \sin(\boldw^{\T}\boldz) \boldB(\boldw)\boldB(\boldw)^\T p(\boldw) d\boldw
\end{align}
where the trigonometric identity ${\cos(\alpha \pm \beta) = \cos\alpha \cos\beta \mp \sin\alpha \sin\beta}$ is applied. Then \eqref{odd-kernel-expected-value} follows. The approximation \eqref{odd-kernel-rff-approx} is the empirical mean for the sample $\boldw_1,\ldots, \boldw_d $. 
\hboks

The optimal solution of \eqref{eq:vector_valued_regular_least_squares} is $\hat\boldf(\boldx) = \boldPsi_o(\boldx)^{\T} \boldsymbol{\alpha}$  
where the coefficient vector $\boldsymbol{\alpha} \in\R^{dn_1}$ is computed from the ${dn_1}$ dimensional linear system
\begin{equation}\label{eq:rff_odd_vector_valued_regular_least_squares_minimum}
    \left( \sum_{i=1}^N \left(\boldPsi_o(\boldx_i)\boldPsi_o(\boldx_i)^{\T} + \lambda \eye_{dn_1}\right) \right) \boldalpha = \sum_{i=1}^N \boldPsi_o(\boldx_i) \boldy_i
\end{equation}

\textbf{Remark 1:} An even kernel can be defined by $\boldK_{\mathrm{even}}(\boldx,\boldz) = \frac{1}{2} (\boldK(\boldx,\boldz) + \boldK(-\boldx,\boldz))$, and the RFF approximation can be found as in the odd case to be given by $\boldK_{\mathrm{even}}(\boldx,\boldz) \approx \boldPsi_e(\boldx)^{\T} \boldPsi_e(\boldz)$ with
\begin{equation}\label{even-kernel-rff-approx-psi}
    \boldPsi_e(\boldx) = \frac{1}{\sqrt{d}} 
    \begin{bmatrix}
        \cos(\boldw_{1}^{\T}\boldx)\boldB(\boldw_{1})^{\T}\\
        \vdots\\
        \cos(\boldw_{d}^{\T}\boldx)\boldB(\boldw_{d})^{\T}
    \end{bmatrix}
    \in \R^{dn_1 \times n}
\end{equation}

\subsection{Symplectic kernel}

In this section the symplectic kernel is analyzed. This vector-valued reproducing kernel was presented in \cite{Boffi2022}, where it was used with an RFF approximation for nonparametric adaptive prediction for Hamiltonian dynamics. A similar kernel which included dissipation terms was presented with RFF approximation for use in Gaussian processes in \cite{Tanaka2022}. In this section the symplectic kernel is further analyzed, and the relation between the RFF for the vector field and the RFF for the Hamiltonian is established. 


Let $\boldx,\boldz\in\R^n$, and let $\boldG_{c}(\boldx)\in \R^{n\times n}$ be the signature in \eqref{eq:curl_free_kernel_from_gaussian} for the curl-free kernel $\boldK_{c}$.

\textbf{Proposition 2:}
{\em
The shift invariant function $\boldK_{s}(\boldx,\boldz) = \boldG_s(\boldx-\boldz) \in \R^{n\times n}$ defined by signature 
\beq\label{eq:symplectic_kernel_base}
    \boldG_{s}(\boldx) = \boldJ \boldG_{c}(\boldx) \boldJ^{\T}
\eeq  
is a vector-valued reproducing kernel which defines an RKHS $\calH_s$ of functions $\boldf\in \calH_s$ so that $\dot\boldx = \boldf(\boldx)$ is Hamiltonian.

}

Proof: The curl-free kernel $\boldK_c(\boldx,\boldz) = \boldG_c(\boldx-\boldz)$ defined in \eqref{eq:curl_free_kernel_from_gaussian} is positive definite since it is a reproducing kernel. The signature satisfies 
\beq
\boldG_c(\boldJ\boldx)  
= \frac{1}{2\sigma^2}e^{-\frac{\boldx^\T\boldJ^\T\boldJ\boldx}{2\sigma^2}} 
\left( \eye - \frac{\boldJ\boldx\boldx^\T\boldJ^\T}{\sigma^2}  \right)
= \boldJ \boldG_{c}(\boldx) \boldJ^{\T}
\eeq
where it is used that $\boldJ\boldJ^\T = \boldJ^\T\boldJ = \eye$. It follows that $\boldG_s(\boldx) = \boldG_c(\boldJ\boldx)$. The symplectic kernel $\boldK_{s}(\boldx,\boldz)$ is therefore positive definite since 
\beq\label{Ks(x,z)=K_c(Jx,Jz)}
\boldK_s(\boldx,\boldz) = \boldK_c(\boldJ\boldx,\boldJ\boldz)
\eeq
The function value of $\boldf\in \calH_s$ is given by 
\begin{align}
\boldf(\boldx) &= \sum_{i=1}^N \boldK_s(\boldx,\boldx_i)\bolda_i\\
&= -\sum_{i=1}^N \boldJ \boldnabla \boldnabla^{\T}k_\sigma(\boldx,\boldx_i)\boldJ^{\T} \bolda_i \\
&= -\boldJ \boldnabla \sum_{i=1}^N \boldnabla^{\T}k_\sigma(\boldx,\boldx_i) \boldc_i
\end{align}
where ${\boldc_{i} = \boldJ^{\T} \bolda_{i}}$ and differentiation is with respect to $\boldx$. This results in the Hamiltonian dynamics $\boldf(\boldx) = \boldJ \boldnabla H(\boldx)$ where the Hamiltonian is
\beq\label{eq:learned_hamiltonian}
    {H}(\boldx) = -\sum_{i = 1}^{N} \boldnabla^{\T} k_\sigma(\boldx,\boldx_i) \boldc_{i}
\eeq
It follows that the system is Hamiltonian. 
\hboks


The RFF approximation for the symplectic kernel is found from \eqref{RFF-K-general-case} and \eqref{RFF-general-case} with $\boldB(\boldw) = \boldJ\boldw$ and ${\boldw \sim p_{\sigma}(\boldw)}$.

\subsection{RFF approximation for the odd symplectic kernel}


The odd symplectic kernel 
\beq\label{eq:odd_symplectic_kernel}
\boldK_{s,\mathrm{odd}}(\boldx,\boldz) = \frac{1}{2} (\boldK_s(\boldx,\boldz) - \boldK_s(-\boldx,\boldz))
\eeq was defined in \cite{Smith2024} by applying \eqref{eq:odd_kernel} to the symplectic kernel $\boldK_s$. It follows from Lemma 1 that any function $\boldf = \sum_{i=1}^N \boldK_{s,\mathrm{odd}}(\cdot,\boldx_i) \bolda_i \in \calH_{s,\mathrm{odd}}$ where $\calH_{s,\mathrm{odd}}$ is the RKHS defined by $\boldK_{s,\mathrm{odd}}$ will then be odd, and $\dot\boldx = \boldf(\boldx)$ will be Hamiltonian. 

In this section the RFF approximation of the kernel is derived. The RFF approximation of the odd symplectic kernel is found from \eqref{odd-kernel-rff-approx} 
with 
\begin{equation}\label{eq:odd_symplectic_kernel_rff}
    \boldPsi_{s,o}(\boldx) = \frac{1}{\sqrt{d}} 
    \begin{bmatrix}
        \sin(\boldw_{1}^{\T}\boldx)(\boldJ\boldw_{1})^{\T}\\
        \vdots\\
        \sin(\boldw_{d}^{\T}\boldx)(\boldJ\boldw_{d})^{\T}
    \end{bmatrix}
    \in \R^{d \times n}
\end{equation}
The function value of the vector field is then 
\beq
\hat \boldf(\boldx) = \boldPsi_{s,o}(\boldx) \boldalpha
\eeq
Following \cite[Equation 8]{Szabo2019}, the corresponding approximation for the Hamiltonian is set to 
\begin{equation}\label{eq:learned_hamiltonian_rff_vector}
    \hat{H}(\boldx) = \boldGamma(\boldx)^{\T} \boldalpha
\end{equation}
where
\begin{equation}
    \boldGamma(\boldx) = \frac{1}{\sqrt{d}} 
    \begin{bmatrix}
        \cos(\boldw_{1}^{\T}\boldx)\\
        \vdots\\
        \cos(\boldw_{d}^{\T}\boldx)
    \end{bmatrix}
    \in \R^{d}
\end{equation}
since this gives $\hat \boldf(\boldx) = \boldJ\boldnabla\hat H(\boldx)$. It is noted that an odd vector field $\boldf$ corresponds to an even Hamiltonian $H$. 



\section{Experiments}\label{sec:5_experiments}

The proposed kernel was evaluated in simulations where the Hamiltonian dynamics of three Hamiltonian systems with odd vector fields were learned. The RFF approximation of the Gaussian separable kernel \eqref{Gaussian-separable-kernel} and the RFF approximation of the odd symplectic kernel in \eqref{eq:odd_symplectic_kernel_rff} were used and compared for the three systems. For all experiments, trajectories were generated by simulating the true system using a Runge–Kutta 89 integrator \cite{Verner2010} \textit{ode89} in MATLAB. The empirical mean square error (MSE) is used both in the tuning of the hyperparameters ${\sigma}$ and ${\lambda}$ and for reporting the simulation results. MSE is calculated for ${N}$ number of trajectories of time duration ${T}$ as
\begin{equation}
    \text{MSE} = \frac{1}{N} \sum_{i=1}^{N} \frac{1}{T_i} \sum_{t=0}^{T_i} \| \boldx_{t,i} - \boldxhat_{t,i} \|_{2}^{2}
\end{equation}
where ${\boldx_{t,i} \in \Rn}$ is the true state and ${\boldxhat_{t,i} \in \Rn}$ is the learned system state.

\subsection{Hyperparameter tuning}

The hyperparameters ${\sigma}$ and ${\lambda}$ greatly influence the learned vector fields. The hyperparameters were tuned using a genetic algorithm \cite{Goldberg1989} in MATLAB, by minimizing the cross-validation error \cite{Kohavi1995} over the training set, with the following bounds applied to the hyperparameters: ${\sigma \in \bb 1, 30 \eb}$ and ${\lambda \in \bb 10^{-8}, 10^{-1} \eb}$. The training set ${\calZ = \{ (\boldx_{i},\boldy_{i}) \in \Rn \times \Rn \}_{i=1}^{N}}$ was split into mutually exclusive subsets ${\calZ_1, \dots \calZ_k}$, and for each iteration ${i \in \{ 1, \dots, k \}}$, the model was trained on the subset ${\hat{\calZ}_{i} = \calZ \setminus \calZ_i}$ and evaluated on ${\calZ_i}$. Formally, the hyperparameter optimization is written as \cite{Krejnik2012}
\begin{equation}
    \min_{\sigma,\lambda} \frac{1}{k} \sum_{i = 1}^{k} \text{MSE}\left( \boldf_{\hat{\calZ}_{i}}, \calZ_i \right)
\end{equation}
where ${\boldf_{\hat{\calZ}_{i}}}$ is the learned vector field trained on ${\hat{\calZ}_{i} = \calZ \setminus \calZ_i}$, using the hyperparameters ${\sigma}$ and ${\lambda}$, and MSE is the empirical mean square error between the learned model and ${ \calZ_i}$.

\subsection{Simple pendulum}\label{sec:simple_pendulum}

\begin{figure*}[hbt!]
    \centering
    \begin{subfigure}[h]{0.32\textwidth}
    \centering
        \includegraphics[width=\textwidth]{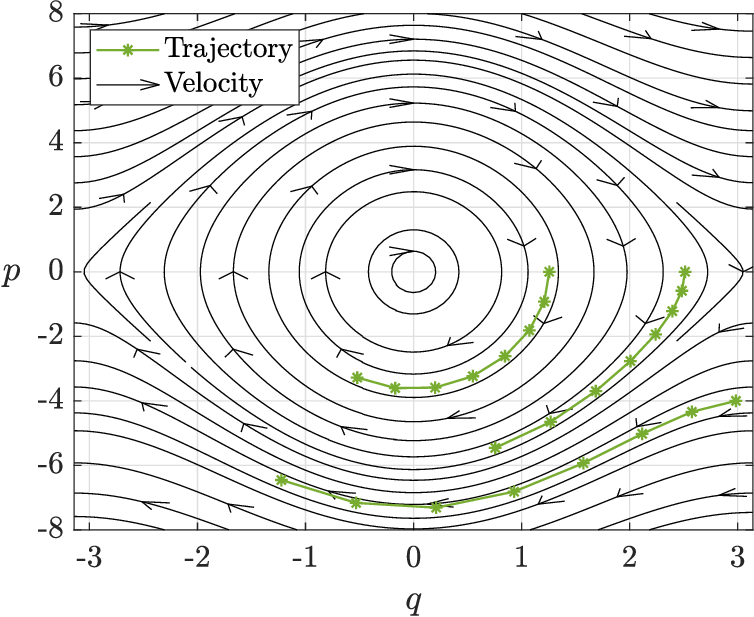}
        \caption{True system}
        \label{fig:simple_pendulum_base_model}
    \end{subfigure}
    \hfill
    \begin{subfigure}[h]{0.32\textwidth}
    \centering
        \includegraphics[width=\textwidth]{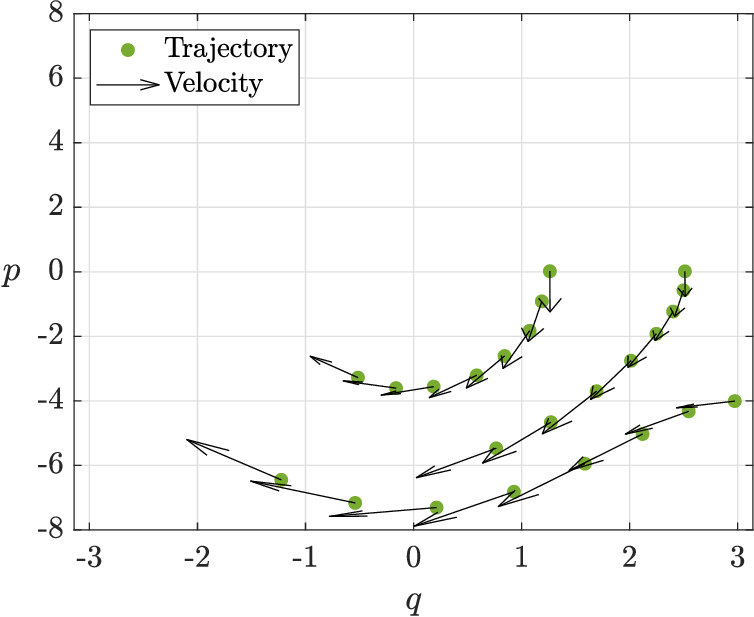}
        \caption{Data set}
        \label{fig:simple_pendulum_data_set}
    \end{subfigure}
    \hfill
    \begin{subfigure}[h]{0.32\textwidth}
    \centering
        \includegraphics[width=\textwidth]{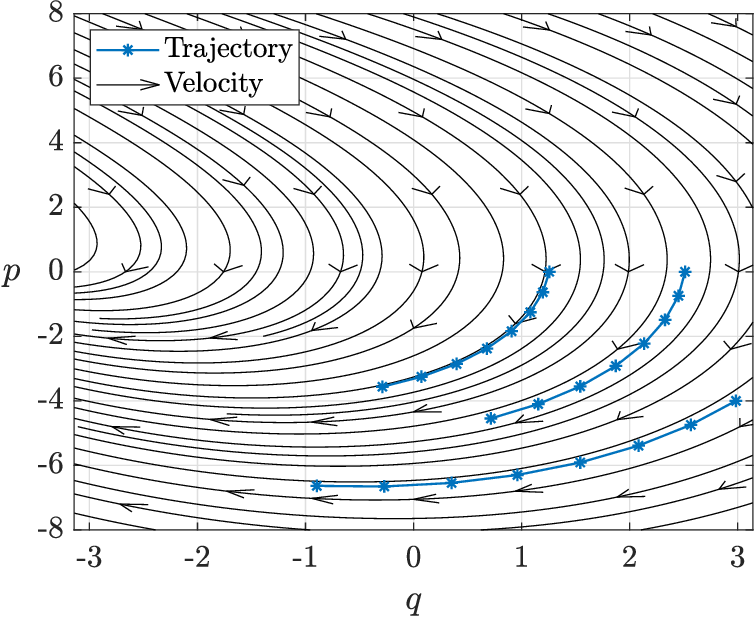}
        \caption{Gaussian model}
        \label{fig:simple_pendulum_gaussian_kernel}
    \end{subfigure}\\
    \vspace{2mm}
    \begin{subfigure}[h]{0.32\textwidth}
    \centering
        \includegraphics[width=\textwidth]{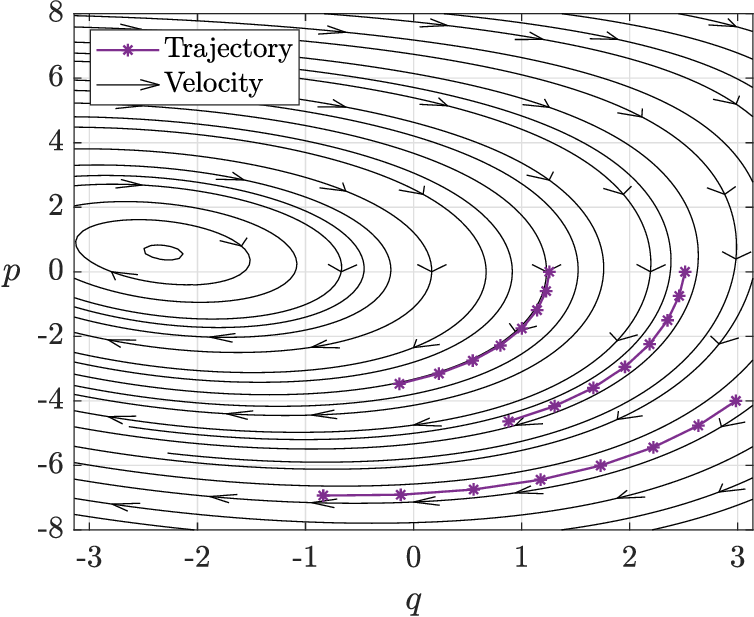}
        \caption{Symplectic model}
        \label{fig:simple_pendulum_symplectic_kernel}
    \end{subfigure}
    \hfill
    \begin{subfigure}[h]{0.32\textwidth}
    \centering
        \includegraphics[width=\textwidth]{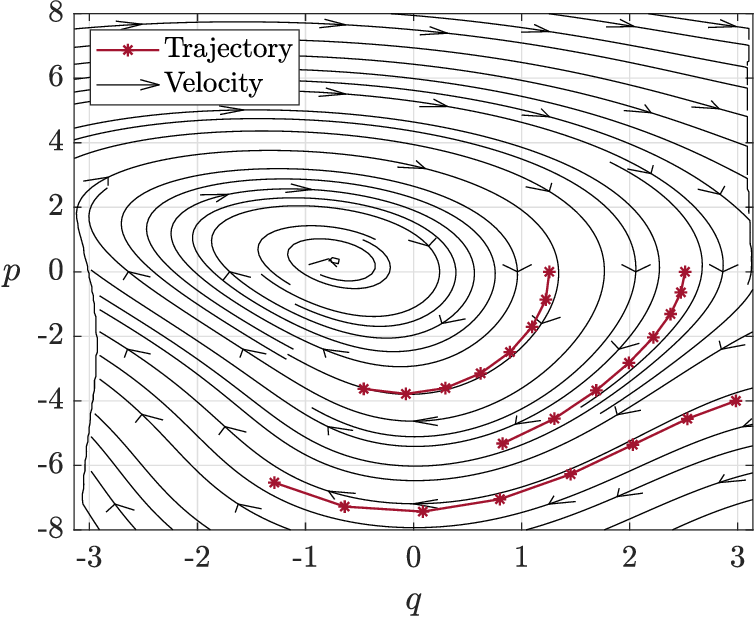}
        \caption{SympGPR model}
        \label{fig:simple_pendulum_symp_gpr}
    \end{subfigure}
    \hfill
    \begin{subfigure}[h]{0.32\textwidth}
    \centering
        \includegraphics[width=\textwidth]{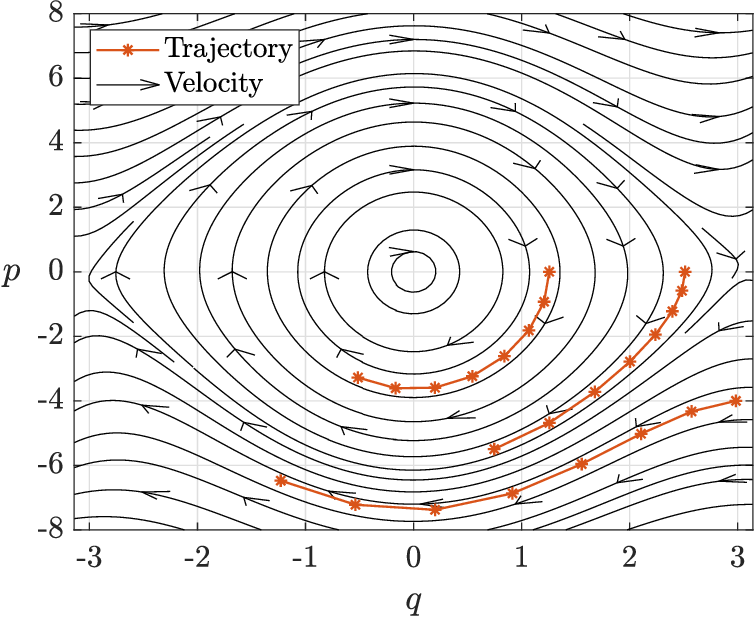}
        \caption{Odd sympl. model}
        \label{fig:simple_pendulum_odd_symplectic_kernel}
    \end{subfigure}
    \caption{Stream and trajectory plots for the simple pendulum and extracted data set, and the resulting learned models using the separable Gaussian kernel, symplectic kernel, SympGPR, and the odd symplectic kernel.}
    \label{fig:simple_pendulum}
\end{figure*}

A simple pendulum is modeled with a point mass ${m}$ at the end of a mass-less rod of length ${l}$. The pendulum angle is ${\theta}$. The equation of motion is given by
\beq
    \thetaddot = -\frac{g}{l} \sin(\theta)
\eeq
where ${g}$ is the acceleration of gravity. The generalized coordinate is ${q = \theta}$, the kinetic energy is ${T = \frac{1}{2} m l^2 \qdot^2}$ and the potential energy is ${U = m g l (1 - \cos(q))}$. The generalized momentum is ${p = m l^2 \qdot}$. The Hamiltonian is
\begin{equation}\label{eq:simple_pendulum_hamiltonian}
    H(q,p) = \frac{p^2}{2 m l^2} + m g l (1 - \cos(q))
\end{equation}
The Hamiltonian dynamics are then given by
\begin{equation}\label{eq:simple_pendulum_hamiltonian_dynamics}
    \qdot = \dfrac{\partial H}{\partial p} = \frac{p}{m l^2}, \qquad \pdot = -\dfrac{\partial H}{\partial q} = - m g l \sin(q)
\end{equation}
Figure~\ref{fig:simple_pendulum_base_model} shows the true system with parameters ${m = 1}$, ${l = 1}$, and ${g = 9.81}$. Three trajectories were generated, and the system was simulated with three different initial conditions: ${\boldx_{0} = \{ \bb \frac{2\pi}{5}, 0\eb^{\T} , \bb \frac{4\pi}{5}, 0\eb^{\T} , \bb \frac{19\pi}{20}, -4 \eb^{\T} \}}$. The time step was set to ${h = 0.1}$ as the system was simulated for ${t \in \bb 0, 0.7 \eb}$ seconds, giving ${8}$ data points for each trajectory, and ${N = 24}$ total data points. The velocities ${\boldy}$ were sampled at each trajectory point, and zero mean Gaussian noise with standard deviation ${\sigma_n = 0.01}$ was added to the trajectory and velocity data. Noise was added also to $\boldx$ to make the simulations closer to a realistic experimental setting. Figure~\ref{fig:simple_pendulum_data_set} shows the resulting data set.

The ${d = 50}$ random features were used for the Gaussian model, and ${d = 400}$ random features were used for the odd symplectic model, giving an equal number of model coefficients ${\boldalpha}$ for each model. For additional comparison, the symplectic kernel proposed in {\cite{Boffi2022}} and the SympGPR method proposed in {\cite{Rath2021}} were also used to learn the dynamics of the simple pendulum. ${d = 200}$ random features were used for the symplectic kernel. A smaller time step of ${h = 0.025}$ was used for generating the training data for the SympGPR model to achieve satisfactory results when plotting the phase plot.

\begin{figure}[ht!]
    \centering
    \begin{subfigure}[h]{\columnwidth}
    \centering
        \includegraphics[width=\textwidth]{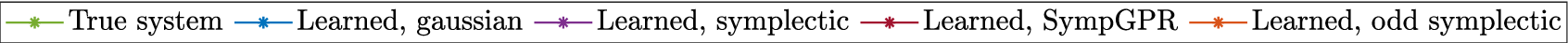}
    \end{subfigure}\\
    \vspace{2mm}
    \begin{subfigure}[h]{0.32\columnwidth}
    \centering
        \includegraphics[width=\textwidth]{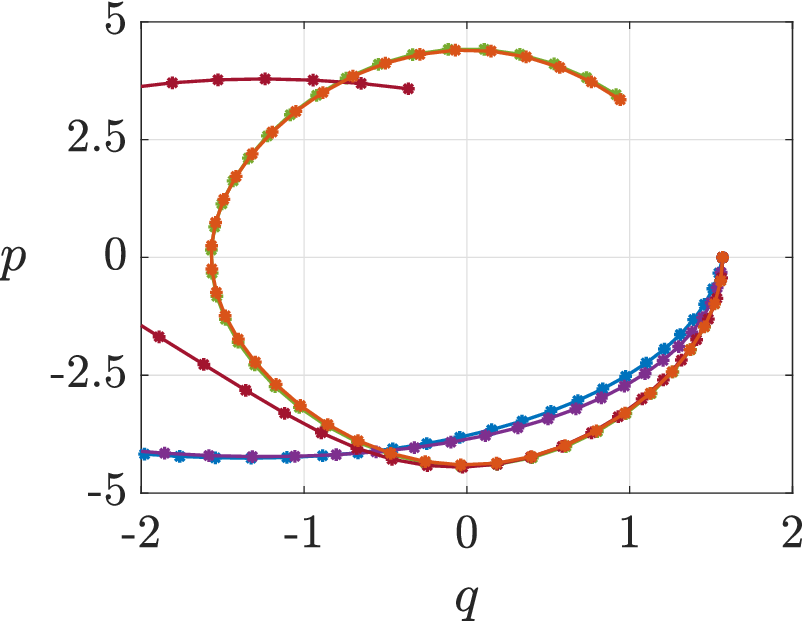}
        \caption{Test trajectory}
        \label{fig:simple_pendulum_test_trajectory}
    \end{subfigure}
    \hspace{5mm}
    \begin{subfigure}[h]{0.6\columnwidth}
    \centering
        \includegraphics[width=\textwidth]{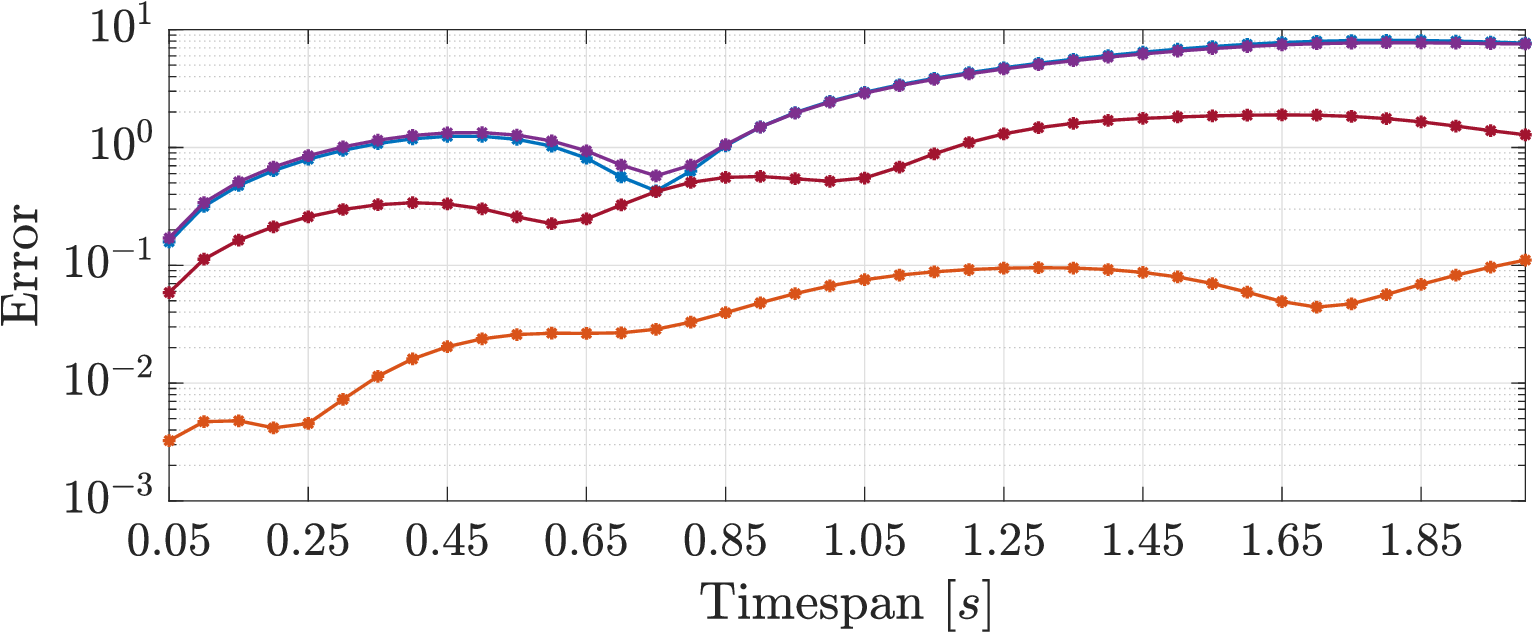}
        \caption{Test trajectory error}
        \label{fig:simple_pendulum_test_trajectory_error}
    \end{subfigure}
    \caption{Comparison of the four learned models against the simple pendulum system, using the test trajectory.}
    \label{fig:simple_pendulum_test}
\end{figure}

Figures~\ref{fig:simple_pendulum_gaussian_kernel} and \ref{fig:simple_pendulum_odd_symplectic_kernel} show phase plots of the learned models using the separable Gaussian kernel and the odd symplectic kernel, respectively. The function learned with the Gaussian separable kernel did not accurately represent the true dynamics from such a limited dataset. The model learned with the odd symplectic kernel was accurate and gave a good representation of the vector field of the simple pendulum system. It is seen from Figure~\ref{fig:simple_pendulum_odd_symplectic_kernel} that symmetry and energy conservation lead to periodic orbits like the true system. The symplectic kernel also failed to learn the dynamics of the simple pendulum accurately (Figure~\ref{fig:simple_pendulum_symplectic_kernel}). The vector field is similar to the learned Gaussian model, but the symplectic kernel enforced energy conservation evident from the periodic orbits. The SympGPR method reproduces the true system's vector field close to the training set but fails to generalize to the entire phase plot as shown in Figure~{\ref{fig:simple_pendulum_symp_gpr}}. The lack of periodic orbits might be due to the use of numerical differentiation to get the derivative information for the streams.

A separate test trajectory was simulated to test the generalized performance of the learned models. The initial condition was ${\boldx_{0} = \bb \frac{\pi}{2}, 0\eb^{\T}}$ and the time horizon is ${t \in \bb 0, 2 \eb}$ seconds. The error between the true system and the learned model trajectories was defined as ${\text{Err} = \| \boldx_{gt} - \boldx_{l} \| }$. Figure~\ref{fig:simple_pendulum_test_trajectory} shows the five resulting trajectories, and Figure~\ref{fig:simple_pendulum_test_trajectory_error} shows the error for each time step. The results showed that the odd symplectic model was far more accurate than the Gaussian separable, symplectic, and SympGPR models, as all failed to generalize beyond the area close to the data set.

\subsection{Cart-pole}

The Cart-Pole \cite{Saber2001} is a planar, underactuated mechanical system where the task is to balance an inverted pendulum starting at an arbitrary initial condition, using only the linear motion of the cart as the input. For this task of system identification, the un-actuated system is modeled. The system consists of a cart moving linearly in the horizontal ${x}$-direction with mass ${m_{c}}$ and an inverted pendulum with point mass ${m_{p}}$ and massless rod with length ${l}$, connected to the cart through a pivot. The angle between the pendulum and the vertical axis is denoted by ${\theta}$, which is zero at the upright position. The kinematics of the system are given by
\begin{equation}
    \boldr_{c} = \bb x \\ 0 \eb, \quad \boldr_{p} = \bb x + l \sin(\theta) \\ l \cos(\theta) \eb
\end{equation}
where ${\boldr_{c}}$ and ${\boldr_{p}}$ are the positions in the ${xy}$-plane of the cart ${m_{c}}$ and pendulum ${m_{p}}$, respectively. The velocities are
\begin{equation}
    \boldv_{c} = \bb \xdot \\ 0 \eb, \quad \boldv_{p} = \bb \xdot + l \thetadot \cos(\theta) \\ -l \thetadot \sin(\theta) \eb
\end{equation}
The kinetic energy ${T}$ and potential energy ${U}$ of the system are
\begin{align}
    T   &= \frac{1}{2} \left( m_{c} + m_{p} \right) \xdot^2 + m_{p} l \xdot \thetadot \cos(\theta) + \frac{1}{2} m_{p} l^{2} \thetadot^{2}\\
    U   &= m_{p} g l \cos(\theta)
 \end{align}
Defining the generalized coordinate ${\boldq = \bb x, \theta \eb^{\T}}$ and its time derivative ${\boldqdot = \bb \xdot, \thetadot \eb^{\T}}$, the generalized momentum is ${\boldp = M(\boldq) \boldqdot}$ where the mass matrix is
\begin{equation}
    M(\boldq) = \bb ( m_{c} + m_{p}) & m_{p} l \cos(\theta) \\ m_{p} l \cos(\theta) & m_{p} l^{2} \eb
\end{equation}
The Hamiltonian of the system is
\begin{equation}\label{eq:cart_pole_hamiltonian}
    H(\boldq,\boldp) = \frac{1}{2} \boldp^{\T} M(\boldq)^{-1} \boldp + U(\boldq)
\end{equation}
where
\begin{equation}
    U(\boldq) = m_{p} g l \cos(\theta)
\end{equation}
Finally, the Hamiltonian dynamics are written as
\begin{align}\label{eq:cartpole_hamiltonian_dyn}
    \boldqdot &= \dfrac{\partial H}{\partial \boldp} = M(\boldq)^{-1} \boldp\\
    \boldpdot &= -\dfrac{\partial H}{\partial \boldq} = - \left( \frac{1}{2} \boldp^{\T} \dfrac{\partial M(\boldq)^{-1}}{\partial \boldq} \boldp + \dfrac{\partial U(\boldq)}{\partial \boldq} \right)
\end{align}

The parameters of the true system were ${m_c = 0.8}$, ${m_p = 0.5}$, ${l = 1}$, and ${g = 9.81}$. The training set was generated by uniformly sampling an increasing number of initial conditions in the interval
\begin{equation}
    \bb -2 \\ -\pi \\ -2 \\ -2 \eb
    \leq \bb q_1 \\ q_2 \\ p_1 \\ p_2 \eb \leq 
    \bb 2 \\ \pi \\ 2 \\ 2 \eb
\end{equation}
The number of initial conditions was 15, 31, 63, 127, 255, 511, and 1023. For each initial condition, the true system was simulated for ${t \in \bb 0, 2 \eb}$ seconds, with 30 times steps in each trajectory. The velocities ${\boldy}$ were sampled at each trajectory point, and zero mean Gaussian noise with standard deviation ${\sigma_n = 0.01}$ was added to the trajectory and velocity data. A separate test set was generated by uniformly sampling 10 initial conditions in the interval
\begin{equation}
    \bb -2 \\ -\pi \\ -2 \\ -2 \eb
    < \bb q_1 \\ q_2 \\ p_1 \\ p_2 \eb <
    \bb 2 \\ \pi \\ 2 \\ 2 \eb
\end{equation}
and simulating the true system for ${t \in \bb 0, 2 \eb}$ seconds, with 30 times steps in each trajectory. The experiments were conducted 20 times for each number of initial conditions by resampling the training set and test set for each run.

The ${d = 50}$ random features were used for the Gaussian model, and ${d = 400}$ random features were used for the odd symplectic model, giving an equal number of model coefficients ${\boldalpha}$ for each model.

\begin{figure*}[hbt!]
    \centering
    \begin{subfigure}[h]{0.5\columnwidth}
    \centering
        \includegraphics[width=\textwidth]{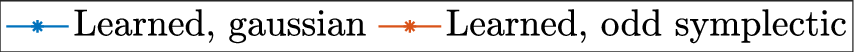}
    \end{subfigure}\\
    \vspace{2mm}
    \centering
    \begin{subfigure}[h]{0.49\textwidth}
    \centering
        \includegraphics[width=\textwidth]{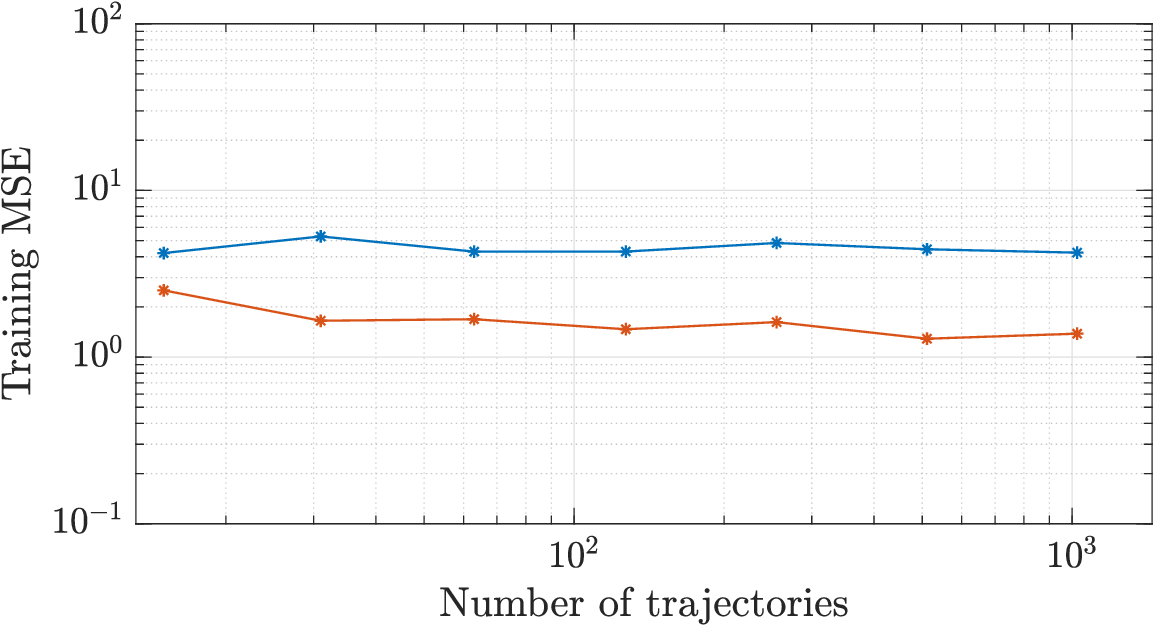}
        \caption{Training MSE}
        \label{fig:learning_w_side_info_cart_pole_training_mse}
    \end{subfigure}
    \hfill
    \begin{subfigure}[h]{0.49\textwidth}
    \centering
        \includegraphics[width=\textwidth]{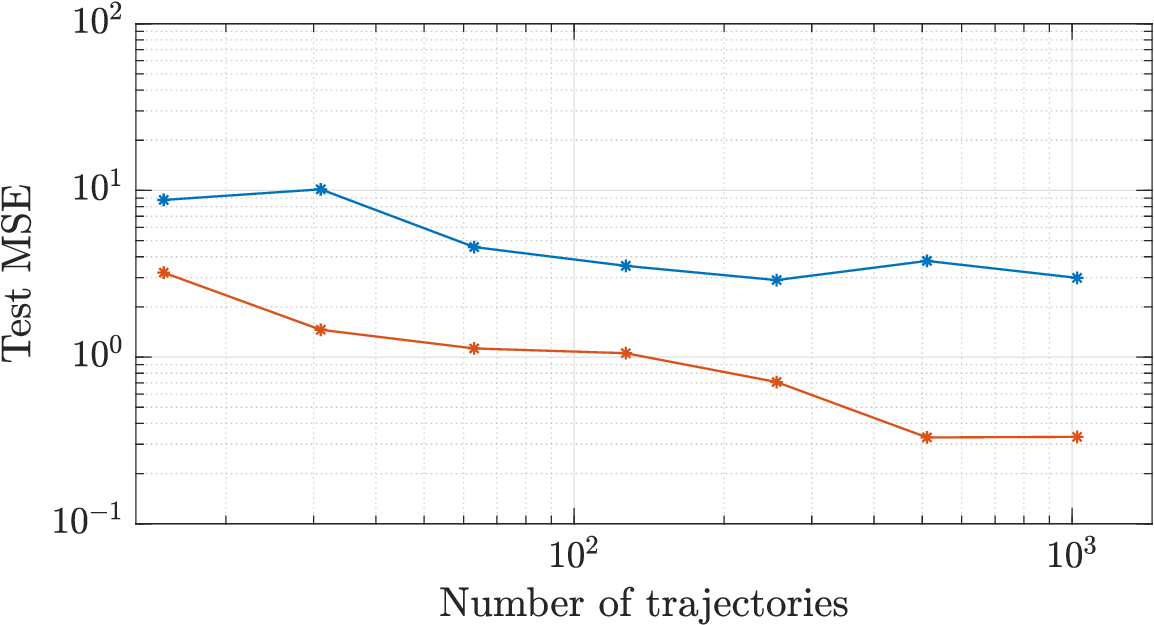}
        \caption{Test MSE}
        \label{fig:learning_w_side_info_cart_pole_test_mse}
    \end{subfigure}
    \caption{Cart-pole: Mean MSE for the training set and test set over 20 different seeds for each number of initial conditions in the training set. Axes are in log-log scale}
    \label{fig:cart_pole_mse}
\end{figure*}

The final learned models were simulated using the same initial conditions and time horizon as the true system, and the resulting trajectories were compared by calculating the MSE for both the training trajectories and the test trajectories. 

The experiments showed that the odd symplectic model outperformed the Gaussian model for both the training set and the test set. Showing an improvement in both accuracy and generalizability. Beyond a consistently lower mean MSE for both the training and test sets, the odd symplectic model outperforms the Gaussian model with fewer training trajectories. For the training set, the odd symplectic model trained on just 15 trajectories outperforms the Gaussian model for every number of trajectories. For the test set, the Gaussian model requires 255 trajectories in the training set to match the performance of the odd symplectic model trained on just 15 trajectories. This can be observed in Figure \ref{fig:cart_pole_mse}, where the mean training MSE and mean test MSE are shown for each number of initial conditions in the training set.

\subsection{Two-link Planar Robot}

The two-link planar robot \cite{Saber2001}, also known as Acrobot or Pendubot when underactuated, consists of two pendulums linked together. The first link with uniformly distributed mass ${m_1}$ and length ${L_1}$ rotates about some fixed point with angle ${\theta_1}$ like a simple pendulum. The second link with uniformly distributed mass ${m_2}$ and length ${L_2}$ rotates like a simple pendulum about the end of link 1 with the angle ${\theta_2}$. The Hamiltonian dynamics are derived for the unactuated system for this task of system identification. The zero configuration ${\theta_1 = \theta_2 = 0}$ is for both links to point directly down. The center of masses of the two links are at the lengths ${l_1}$ and ${l_2}$ from their respective pivot points. In the inertial frame, the positions of the two center of mass points are given by the kinematics
\begin{equation}
    \boldr_{1} = \bb l_1 \sin \theta_1 \\ -l_1 \cos \theta_1 \eb, \quad \boldr_{2} = \bb l_2 \sin\left(\theta_1 + \theta_2 \right) + L_1 \sin \theta_1 \\ -l_2 \cos\left(\theta_1 + \theta_2 \right) - L_1 \cos \theta_1 \eb
\end{equation}
where ${\boldr_{1}}$ and ${\boldr_{2}}$ are the positions in the ${xy}$-plane of the two idealized masses ${m_{1}}$ and ${m_{1}}$, respectively. The velocities are then
\begin{align}
    \boldv_{1} &= \bb l_1 \omega_1 \cos \theta_1 \\ l_1 \omega_1 \sin \theta_1 \eb\\
    \boldv_{2} &= \bb l_2 \omega_1  \cos\left(\theta_1 + \theta_2 \right) + l_2 \omega_2 \cos\left(\theta_1 + \theta_2 \right) + L_1 \omega_1 \cos \theta_1 \\ l_2 \omega_1 \sin\left(\theta_1 + \theta_2 \right) + l_2 \omega_2 \sin\left(\theta_1 + \theta_2 \right) + L_1 \omega_1 \sin \theta_1 \eb
\end{align}
Using the mass moment of inertia about the center of mass for a slender rod given as ${I = \frac{1}{12} m L^2}$, the kinetic energy ${T}$ of the system is
\begin{equation}
    T = T_1 + T_2
\end{equation}
where
\begin{align}
    T_1 &= \frac{1}{2} \left( m_{1} l_{1}^{2} + I_{1} \right) \omega_{1}^{2}\\
    \begin{split}
         T_2 &= \frac{1}{2} m_{2} ( l_2^2 \omega_1^2 + 2 l_2^2 \omega_1 \omega_2 + 2 l_2 L_1 \omega_1^2 \cos(\theta_2) + l_2^2 \omega_2^2\\
         &\quad+ 2 l_2 L_1 \omega_1 \omega_2 \cos(\theta_2) + L_1^2 \omega_1^2) + \frac{1}{2} I_{2} (\omega_{1}^2 + 2\omega_{1}\omega_{2} + \omega_{2}^{2})
    \end{split}
\end{align}
The potential energy ${U}$ of the system is then derived using the kinematics in the vertical direction
\begin{equation}
    U = g \left( - \left( m_{1} l_{1} + m_{2} L_{1} \right) \cos\left(\theta_{1}\right) - m_{2} l_{2} \cos\left(\theta_{1} + \theta_{2} \right) \right)
\end{equation}
The generalized coordinate is ${\boldq = \bb \theta_{1}, \theta_{2} \eb^{\T}}$ and its time derivative is ${\boldqdot = \bb \omega_{1}, \omega_{2} \eb^{\T}}$. The generalized momentum is ${\boldp = M(\boldq) \boldqdot}$ where the mass matrix is
\begin{equation}
    M(\boldq) = \bb M_{1}  &   M_{2}  \\  M_{2}  &  M_{3}  \eb
\end{equation}
with
\begin{align}
    M&_{1} = m_{1} l_{1}^{2} + m_{2} l_2^2 + m_{2} L_1^2 + I_{1} + I_{2} + 2 m_{2} l_2 L_1 \cos(\theta_2)\\
    M&_{2} = m_{2} l_2^2 + I_{2} + m_{2} l_2 L_1 \cos(\theta_2)\\
    M&_{3} = m_{2} l_2^2 + I_{2}
\end{align}
The Hamiltonian of the system is
\begin{equation}\label{eq:acrobot_hamiltonian}
    H(\boldq,\boldp) = \frac{1}{2} \boldp^{\T} M(\boldq)^{-1} \boldp + U(\boldq)
\end{equation}
where
\begin{equation}
    U(\boldq) = g \left( - \left( m_{1} l_{1} + m_{2} L_{1} \right) \cos\left(\theta_{1}\right) - m_{2} l_{2} \cos\left(\theta_{1} + \theta_{2} \right) \right)
\end{equation}
Finally, the Hamiltonian dynamics are written as
\begin{align}\label{eq:acrobot_hamiltonian_dyn}
    \boldqdot &= \dfrac{\partial H}{\partial \boldp} =  M(\boldq)^{-1} \boldp\\
    \boldpdot &= -\dfrac{\partial H}{\partial \boldq} = - \left( \frac{1}{2} \boldp^{\T} \dfrac{\partial M(\boldq)^{-1}}{\partial \boldq} \boldp + \dfrac{\partial U(\boldq)}{\partial \boldq} \right)
\end{align}

The parameters of the true system were ${m_1 = m_2 = 1}$, ${L_1 = 1}$, ${L_2 = 2}$, ${l_1 = 0.5}$, ${l_2 = 1}$, and ${g = 9.81}$. The training set was generated by uniformly sampling an increasing number of initial conditions in the interval
\begin{equation}
    \bb -\pi \\ -\pi \\ -2 \\ -2 \eb
    \leq \bb q_1 \\ q_2 \\ p_1 \\ p_2 \eb \leq
    \bb \pi \\ \pi \\ 2 \\ 2 \eb
\end{equation}
The number of initial conditions was 15, 31, 63, 127, 255, 511, and 1023. For each initial condition, the true system was simulated for ${t \in \bb 0, 2 \eb}$ seconds, with 30 times steps in each trajectory. The velocities ${\boldy}$ were sampled at each trajectory point, and zero mean Gaussian noise with standard deviation ${\sigma_n = 0.01}$ was added to the trajectory and velocity data. A separate test set was generated by uniformly sampling 10 initial conditions in the interval
\begin{equation}
    \bb -\pi \\ -\pi \\ -2 \\ -2 \eb
    < \bb q_1 \\ q_2 \\ p_1 \\ p_2 \eb <
    \bb \pi \\ \pi \\ 2 \\ 2 \eb
\end{equation}
and simulating the true system for ${t \in \bb 0, 2 \eb}$ seconds, with 30 times steps in each trajectory. The experiments were conducted 20 times for each number of initial conditions by resampling the training set and test set for each run.

The ${d = 100}$ random features were used for the Gaussian model, and ${d = 800}$ random features were used for the odd symplectic model, giving an equal number of model coefficients ${\boldalpha}$ for each model.

The final learned models were simulated using the same initial conditions and time horizon as the true system, and the resulting trajectories were compared by calculating the MSE for both the training trajectories and the test trajectories.

\begin{figure*}[hbt!]
    \centering
    \begin{subfigure}[h]{0.5\columnwidth}
    \centering
        \includegraphics[width=\textwidth]{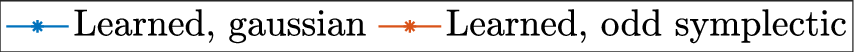}
    \end{subfigure}\\
    \vspace{2mm}
    \centering
    \begin{subfigure}[h]{0.49\textwidth}
    \centering
        \includegraphics[width=\textwidth]{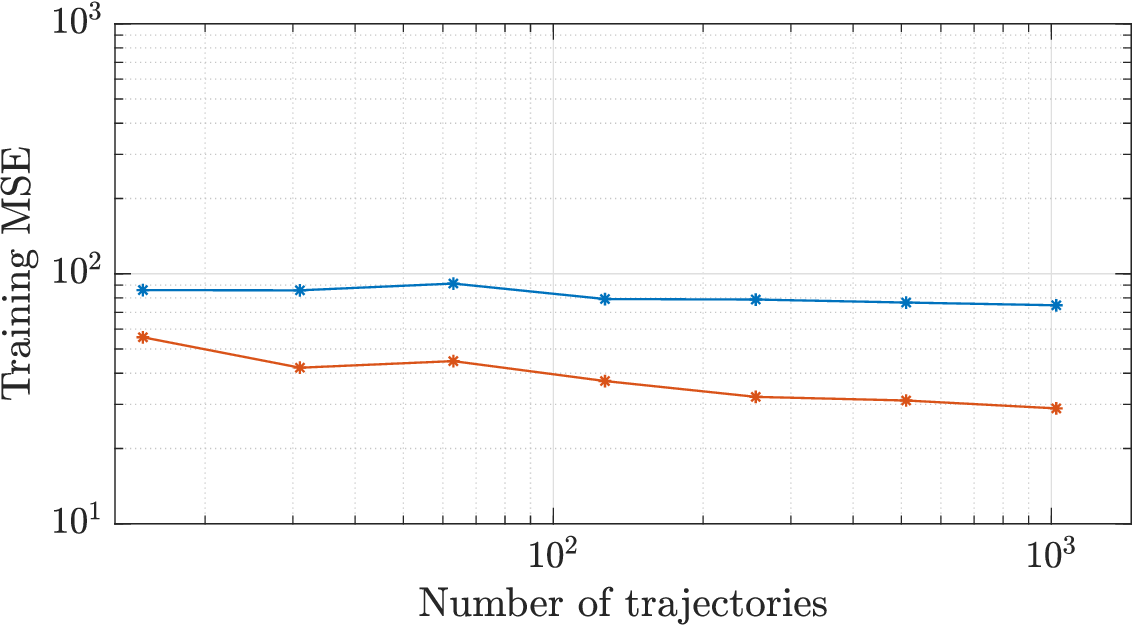}
        \caption{Training MSE}
        \label{fig:learning_w_side_info_acrobot_training_mse}
    \end{subfigure}
    \hfill
    \begin{subfigure}[h]{0.49\textwidth}
    \centering
        \includegraphics[width=\textwidth]{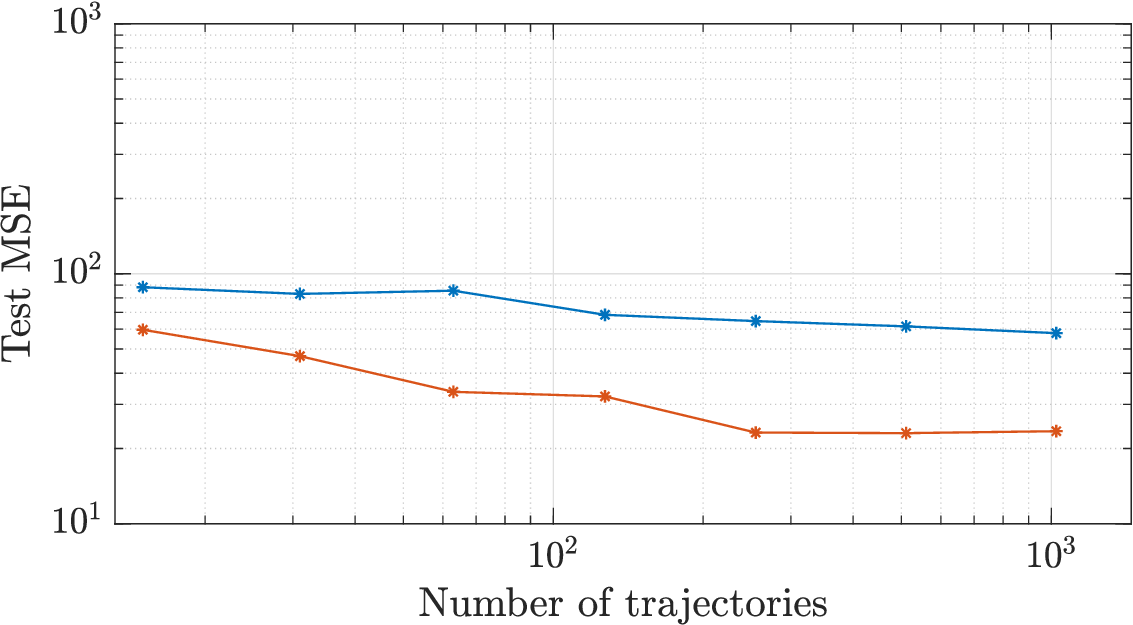}
        \caption{Test MSE}
        \label{fig:learning_w_side_info_acrobot_test_mse}
    \end{subfigure}
    \caption{2-link Robot: Mean MSE for the training set and test set over 20 different seeds for each number of initial conditions in the training set. Axes are in log-log scale}
    \label{fig:acrobot_mse}
\end{figure*}

As with the cart-pole, the odd symplectic model outperforms the Gaussian model with fewer training trajectories. For the training set, the odd symplectic model trained on just 15 trajectories outperforms the Gaussian model across all number of trajectories, and on the test set, the Gaussian model needs 1023 training trajectories to match the odd symplectic model trained on just 15 trajectories. The absolute magnitude of the error is larger for the 2-link robot, which might be due to the chaotic nature of the 2-link robot combined with the noise added to the training data. The results from the experiments can be observed in Figure \ref{fig:acrobot_mse}, where the mean training MSE and mean test MSE are shown for each number of initial conditions in the training set.

\subsection{Varying number of random features}

The odd symplectic kernel was compared to its random feature approximation. The comparison was performed by learning the Hamiltonian dynamics of the simple pendulum given in \eqref{eq:simple_pendulum_hamiltonian_dynamics}, using the odd symplectic kernel \eqref{eq:odd_symplectic_kernel} and its random feature approximation in \eqref{eq:odd_symplectic_kernel_rff}.

The training set was generated by randomly sampling ${5000}$ points in the set ${S = \{ \boldx \in \R^2 \; | \; |q| \leq \pi, |p| \leq 8\}}$. The velocities ${\boldy}$ were sampled at each point, and zero mean Gaussian noise with standard deviation ${\sigma_n = 0.01}$ was added to the trajectory and velocity data. The learned models were evaluated on the three trajectories used as the training set in Section \ref{sec:simple_pendulum}.

The random feature models were learned using an increasing number of random samples ${\boldw}$, and each model was learned using 50 different sets of random samples, using the mean MSE to evaluate the performance.

\begin{figure*}[hbt!]
    \centering
    \begin{subfigure}[h]{0.4\columnwidth}
    \centering
        \includegraphics[width=\textwidth]{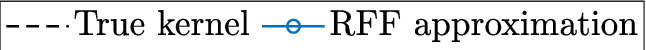}
    \end{subfigure}\\
    \vspace{2mm}
    \centering
    \begin{subfigure}[h]{0.6\textwidth}
    \centering
        \includegraphics[width=\textwidth]{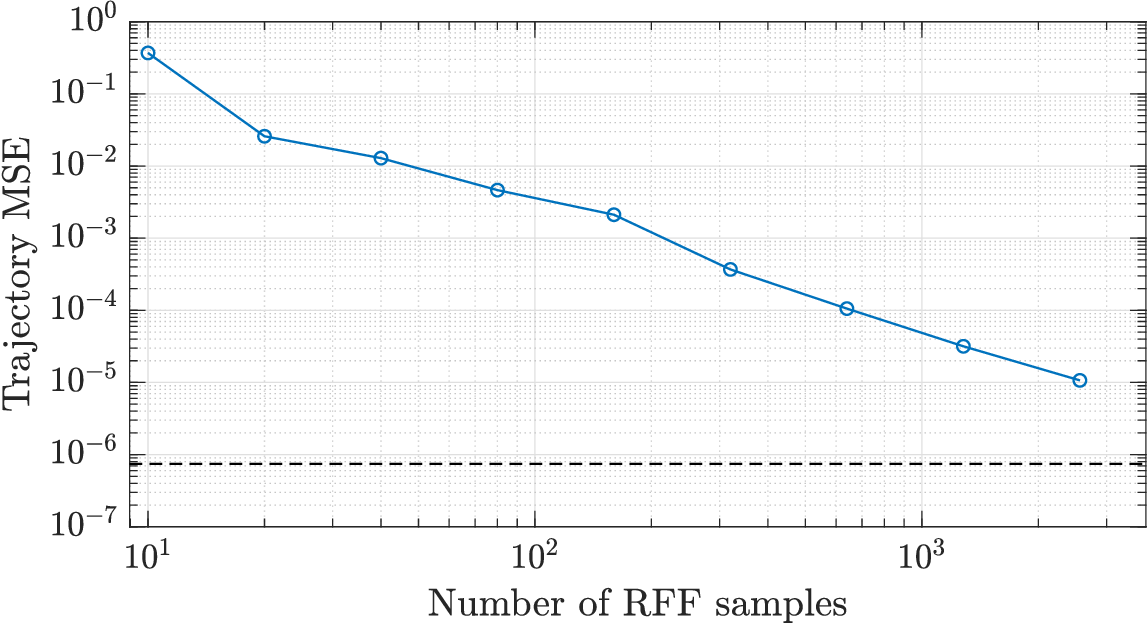}
    \end{subfigure}
    \caption{Trajectory MSE for the odd symplectic kernel and its RFF approximation. The error for the RFF approximation is the mean error over 50 draws of ${d = \{10,20,40,80,160,320,640,1280,2560\}}$ random features.}
    \label{fig:sweep_random_features_loglog}
\end{figure*}

The results show that the true kernel was more accurate than the random feature approximation for the trajectories used in the evaluation, and the error using the random feature decreased exponentially with an increase in the number of random features $d$. The results are shown in Figure \ref{fig:sweep_random_features_loglog}. According to Theorem~12 in \cite{Minh2016}, the random feature approximation will converge exponentially in $d$, and it follows that the approximation of the odd symplectic kernel will converge exponentially in $d$. This result agrees well with the observed exponential convergence of the MSE.

\subsection{Numerical evaluation}

The models were evaluated numerically to investigate the ability of the learned models to capture the side information of the true systems. The odd symmetry was evaluated by sampling ${10\,000}$ points in the right half plane for each of the vector fields, and calculating the odd error given as
\begin{equation}
    e_{\odd} = \| \boldf(\boldx) + \boldf(-\boldx) \|
\end{equation}
where ${\boldf : \Rn \rightarrow \Rn}$ is the dynamical system being evaluated and ${\boldx \in \Rn}$ is the sampled point. As the cart-pole and 2-link robot are learned for different numbers of trajectories, the values corresponding to the maximum mean odd error were used.

\begin{table}[hbt!]
\caption{Odd error ${e_{\odd}}$ for the three dynamical systems}
\centering
\label{tab:oddness_numerical_evaluation}
\footnotesize
\begin{tabular}{@{\extracolsep\fill}lccccccccc}
\toprule
& \multicolumn{2}{c}{Simple pendulum} & \multicolumn{2}{c}{Cart-pole} & \multicolumn{2}{c}{2-link robot}\\
\cmidrule(lr){2-3} \cmidrule(lr){4-5} \cmidrule(lr){6-7}
System & Mean & Variance & Mean & Variance & Mean & Variance\\
\midrule
True ${e_{\odd}}$ & 0.00 & 0.00 & 0.00 & 0.00 & 0.00 & 0.00\\
Gaussian sep. ${e_{\odd}}$ & 7.87 & 6.22 & 2.91 & 1.75 & 14.60 & 34.83\\
Odd symplectic ${e_{\odd}}$ & 0.00 & 0.00 & 0.00 & 0.00 & 0.00 & 0.00\\
\bottomrule
\end{tabular}
\end{table}

The results in Table~\ref{tab:oddness_numerical_evaluation} document that the learned odd symplectic model enforces odd symmetry like the true systems, whereas the Gaussian separable model does not.

The learned Hamiltonian in \eqref{eq:learned_hamiltonian_rff_vector} for the learned odd symplectic models were compared to their corresponding real Hamiltonians in \eqref{eq:simple_pendulum_hamiltonian}, \eqref{eq:cart_pole_hamiltonian}, and \eqref{eq:acrobot_hamiltonian}, over the test trajectories. For the Cart-pole and the 2-link robot, the presented values were selected by selecting for the maximum variance across all test trajectories.

\begin{table}[hbt!]
\caption{Hamiltonian for the three dynamical systems over the test trajectories}
\centering
\label{tab:hamiltonian_numerical_evaluation}
\footnotesize
\begin{tabular}{@{\extracolsep\fill}lcccccc}
\toprule
& \multicolumn{2}{c}{Simple pendulum} & \multicolumn{2}{c}{Cart-pole} & \multicolumn{2}{c}{2-link robot}\\
\cmidrule(lr){2-3} \cmidrule(lr){4-5} \cmidrule(lr){6-7}
Hamiltonian & Mean & Variance & Mean & Variance & Mean & Variance\\
\midrule
Real ${H(\boldx)}$ & $9.81$ & $4.99 \cdot 10^{-15}$ & $3.79$ & $3.09 \cdot 10^{-15}$ & $-0.59$ & $1.44 \cdot 10^{-15}$ \\
Learned ${\hat{H}(\boldx)}$ & $28.93$ & $4.08 \cdot 10^{-15}$ & $5.14$ & $9.97 \cdot 10^{-14}$ & $27.62$ & $5.85 \cdot 10^{-12}$ \\
\bottomrule
\end{tabular}
\end{table}

The results in Table~\ref{tab:hamiltonian_numerical_evaluation} demonstrate that the value of the learned Hamiltonian ${\hat{H}(\boldx)}$ has a constant offset from the true Hamiltonian ${H(\boldx)}$. This agrees with the fact that the potential energy's zero potential cannot be expected to be the same for the learned and true systems. It is seen that the value of the learned Hamiltonian is constant in agreement with \eqref{eq:hamiltonian_time_derivative_zero} since the system is unforced and independent of time. This is reflected in the variance of both ${H(\boldx)}$ and ${\hat{H}(\boldx)}$. Noting that these are results from numerical simulations, the results indicate that the Hamiltonian mechanics are captured in the learned odd symplectic model.

\section{Conclusion}\label{sec:6_conclusion}

A specialized kernel enforcing side information relating to Hamiltonian dynamics and odd symmetry has been presented. The odd symplectic kernel was developed, approximated using random features, and utilized in three comparative experiments. By enforcing the side information through the kernel, the closed-form solution to the learning problem is retained, and the side information is enforced for the whole domain of the learned function. This stands in contrast to the approach of enforcing the side information through the use of constraints in a constrained optimization problem, enforcing the side information only on selected points. Through comparative experiments, we have demonstrated that the proposed kernel outperforms a more standard kernel ridge regression and Gaussian kernel, as the error over both the training set and a separate test set is lower, indicating a more accurate and generalized learned model.

\subsection{Future work}
A challenge with learning Hamiltonian dynamics is the potential lack of data for the generalized momenta and their derivatives. As a result, the method should be extended so that it can be applied using only data for the generalized coordinates and velocities. An alternative is to modify the method using a numerical integrator in the learning procedure as in \cite{Zhong2020}, to eliminate the need for derivative observations. The developed kernel could also form the basis for a GP model to enforce both energy conservation and odd symmetry in a GP model. Furthermore, control-oriented learning could be studied using the proposed kernel.





\bibliographystyle{elsarticle-num} 
\bibliography{utilities/refs}


\end{document}